\pdfoutput=1

\documentclass[11pt]{article}

\PassOptionsToPackage{table}{xcolor}
\usepackage[final]{acl}

\usepackage{times}
\usepackage{latexsym}

\usepackage[T1]{fontenc}

\usepackage[utf8]{inputenc}

\usepackage{microtype}

\usepackage{inconsolata}

\usepackage{graphicx}

\usepackage{tikz}
\usepackage{multirow}
\usepackage{booktabs}
\usepackage{amsmath}
\usepackage{amssymb}

\usepackage[ruled,noend]{algorithm2e}

\usepackage{listings}
\usepackage{colortbl}

\title{Reflective Agreement: Combining Self-Mixture of Agents with a Sequence Tagger for Robust Event Extraction}

\author{
Fatemeh Haji\textsuperscript{1},
Mazal Bethany\textsuperscript{1},
Cho-Yu Jason Chiang\textsuperscript{2},
Anthony Rios\textsuperscript{3},
Peyman Najafirad\textsuperscript{1*} \\
\textsuperscript{1}Secure AI and Autonomy Lab, University of Texas at San Antonio \\
\textsuperscript{2}Peraton Labs \\
\textsuperscript{3}University of Texas at San Antonio \\
\texttt{\{fatemeh.haji, mazal.bethany, anthony.rios, peyman.najafirad\}@utsa.edu} \\
\texttt{jchiang@peratonlabs.com}
}

\begin{document}
\maketitle
\renewcommand{\thefootnote}{\fnsymbol{footnote}}
\footnotetext[1]{Corresponding author.}

\begin{abstract}
\label{abstract}
Event Extraction (EE) involves automatically identifying and extracting structured information about events from unstructured text, including triggers, event types, and arguments.
Traditional discriminative models demonstrate high precision but often exhibit limited recall, particularly for nuanced or infrequent events. Conversely, generative approaches leveraging Large Language Models (LLMs) provide higher semantic flexibility and recall but suffer from hallucinations and inconsistent predictions.
To address these challenges, we propose Agreement-based Reflective Inference System (ARIS), a hybrid approach combining a Self Mixture of Agents with a discriminative sequence tagger. ARIS explicitly leverages structured model consensus, confidence-based filtering, and an LLM  reflective inference module to reliably resolve ambiguities and enhance overall event prediction quality. We further investigate decomposed instruction fine-tuning for enhanced LLM event extraction understanding.
Experiments demonstrate our approach outperforms existing state-of-the-art event extraction methods across three benchmark datasets.
\end{abstract}

\section{Introduction}
\label{introduction}

Event Extraction (EE) aims to identify structured event information from unstructured textual data, including event triggers, event types, and associated arguments with their roles \cite{doddington2004automatic}. Effective event extraction underpins critical applications in information retrieval, knowledge graph construction, and automated decision-making. Despite considerable advancements, robust event extraction remains challenging, primarily due to linguistic variability, semantic complexity, and limited generalization to infrequent or previously unseen events \cite{li2022survey}.

There are two predominant methodologies in EE: discriminative approaches and generative methods leveraging LLMs. Discriminative methods, including transformer-based sequence taggers (e.g., RoBERTa) and structured prediction models, offer superior precision and structural consistency due to their explicit token-level training \cite{zeng-etal-2022-ea2e, liu-etal-2024-beyond-single}. However, these methods often struggle with recall, especially for nuanced or rare events not extensively covered by training datasets. Conversely, generative LLM-based approaches \cite{zhu-etal-2024-lc4ee, gao2024eventrl} demonstrate enhanced semantic flexibility and contextual understanding, achieving broader coverage and improved recall. Yet, these generative approaches frequently produce inconsistent predictions and hallucinations due to their inherent stochasticity, resulting in lower precision in their predictions \cite{meng2024cean}.

Recently, hybrid multi-agent debate-based methods have emerged, leveraging multiple generative LLM agents to iteratively critique and refine predictions \cite{chan2024chateval}. Although promising, these debate approaches have critical limitations: they rely on iterative, often unstructured discussions without explicit grounding, leading to amplified hallucinations and inconsistent outputs; they lack principled mechanisms for systematically resolving persistent disagreements; and they introduce substantial computational overhead with unpredictable inference times. These shortcomings significantly limit their effectiveness and practical applicability.

In this paper, we introduce ARIS (Agreement-based Reflective Inference System), a hybrid event extraction framework explicitly designed to overcome these limitations. ARIS systematically integrates the complementary strengths of a generative Self Mixture-of-Agents, which uses multiple LLM instances decoding in parallel to promote output diversity, with a discriminative sequence tagger that provides essential structural grounding and precision. ARIS introduces the Reflective Agreement mechanism, a structured inference process that explicitly leverages model consensus and confidence-based filtering to select high-confidence event predictions, while employing reflective inference to resolve ambiguities systematically. Crucially, our reflective inference module relies on an LLM explicitly trained to understand the complete event extraction chain (trigger identification, trigger classification, argument identification, and argument classification) through decomposed instruction fine-tuning, significantly enhancing the accuracy and reliability of reflective reasoning.

Our contributions are as follows:
\begin{itemize}
\item We propose ARIS, a hybrid event extraction framework that systematically integrates generative flexibility and discriminative precision, explicitly addressing the limitations inherent to existing debate-based and standalone generative approaches.
\item We introduce Reflective Agreement, a novel structured reflective inference mechanism that leverages explicit model agreement, confidence-based filtering, and contextual reflective reasoning to robustly resolve ambiguous predictions.
\item We demonstrate empirically that ARIS achieves state-of-the-art performance across three event extraction benchmarks, consistently surpassing discriminative, generative, and existing hybrid debate-based methods. Beyond empirical results, ARIS advances theoretical understanding by providing new insights into structured reflective reasoning and hybrid model integration for complex NLP tasks.
\end{itemize}

\section{Related Work}
\label{related_work}
\begin{figure*}[t]
    \centering
    \includegraphics[width=1\linewidth]{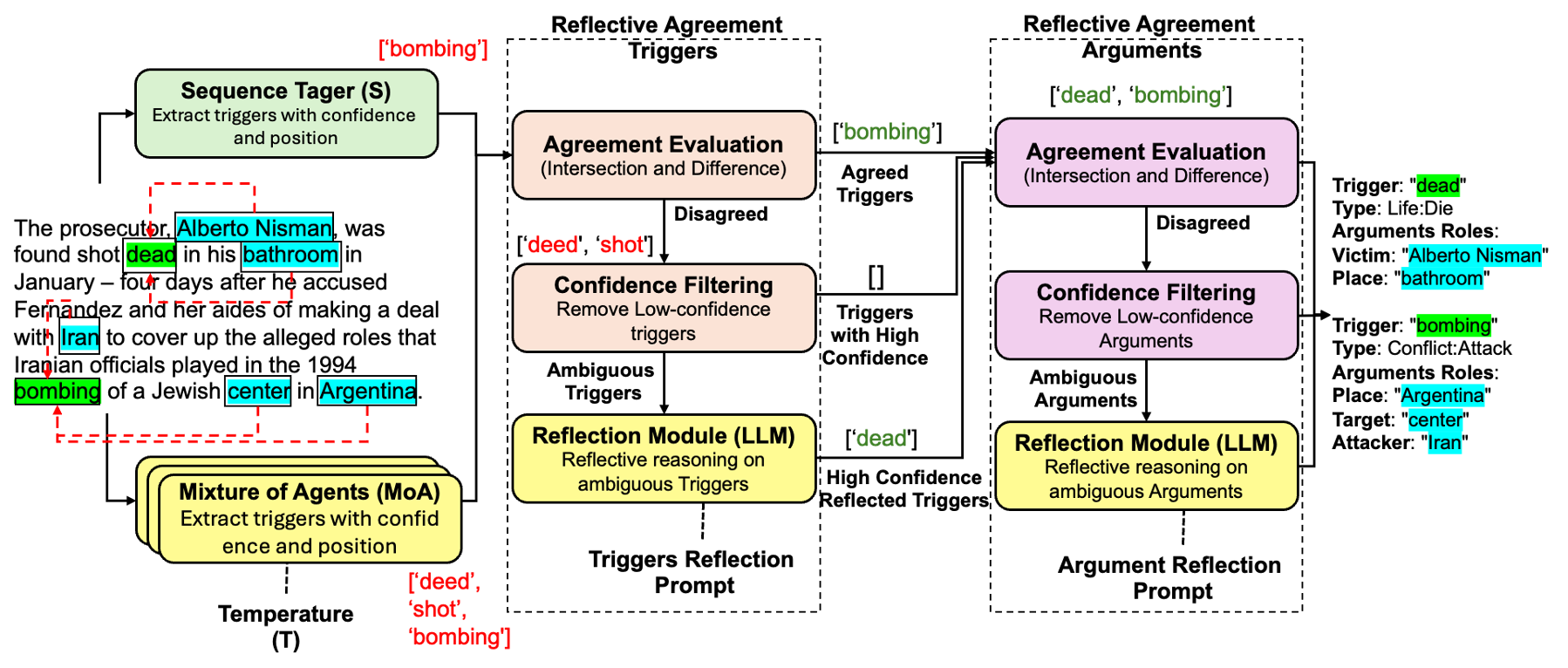}
    \caption{Overview of the proposed ARIS framework illustrating the Reflective Agreement process. ARIS systematically integrates predictions from a discriminative sequence tagger and a generative Self Mixture of Agents. Event triggers and arguments are extracted by each model with associated positions and confidence scores. The framework identifies consented predictions, filters out low-confidence disagreements, and employs a reflective inference module to resolve remaining ambiguities, ultimately producing robust, accurate, and structurally grounded event extraction results.}
    \label{fig:system_figure}
\end{figure*}

\paragraph{Event Extraction with LLMs}

LLMs have emerged as promising tools for Event Extraction tasks, offering strengths in contextual understanding and handling linguistic variation. Several studies have investigated zero-shot and few-shot prompting approaches for Event Extraction with LLMs \cite{chen2024large}. More sophisticated prompting frameworks like the Debate as Optimization (DAO) \cite{wang-huang-2024-debate} employ multiple agent roles to iteratively refine event extraction predictions through structured debate.
Other researchers have explored hybrid approaches combining task-specific models with LLMs. LC4EE \cite{zhu-etal-2024-lc4ee} uses task-specific models for initial Event Extraction, then employs manually defined rules to guide an LLM in verifying and correcting the output. Recent work has begun exploring fine-tuning approaches, with studies incorporating textual descriptions of event types into instruction tuning datasets \cite{srivastava2025instruction} or combining Supervised Fine-Tuning with reinforcement learning \cite{gao2024eventrl}.
However, significant limitations persist across these approaches. Prompting-only methods typically underperform supervised fine-tuning of smaller discriminative models, such as RoBERTa-based models. Hybrid approaches rely on manual rule creation, which limit scalability. Importantly, there is also limited work on systematically designing instruction tuning datasets that address the distinct challenges of event extraction's core subtasks: trigger identification, trigger classification, argument identification, and argument classification. Consequently, many LLM-based methods still fail to outperform supervised fine-tuning of task-specific models.


\paragraph{LLM Instruction Fine-Tuning}

Instruction fine-tuning has emerged as a powerful approach for enhancing language models' capabilities on specific tasks. Recent advancements have focused on structuring the fine-tuning process to improve reasoning abilities and handle complex tasks more effectively.
Chain-of-Thought (CoT) fine-tuning has gained significant attention, where instruction datasets are augmented with reasoning rationales. This enables the models to learn reasoning capabilities \cite{kim2023cot, zelikman2022star, ho2023large}.
Compositional Fine-Tuning addresses complex tasks by explicitly breaking them down into simpler component subtasks \cite{bursztyn2022learning}. Rather than using end-to-end learning, CFT fine-tunes models on a set of component tasks, as well as the end-to-end task, enabling them to learn the end-to-end task more effectively.

\paragraph{LLM Inference Time Improvement}
Advanced inference strategies significantly enhance LLM performance without requiring model parameter updates. Chain-of-Thought prompting \cite{wei2022chain} enables LLMs to break down complex problems into intermediate reasoning steps, improving performance on tasks requiring compositional reasoning. Tree of Thoughts \cite{yao2023tree} extends this approach by allowing models to explore multiple reasoning paths simultaneously, evaluating alternatives and backtracking when necessary. Retrieval-Augmented Generation \cite{lewis2020retrieval} incorporates external knowledge retrieval to improve factuality and reduce hallucination. 
Recent work has also explored self-correction mechanisms for LLMs, where LLMs can iteratively self-correct their outputs through interaction with external tools \cite{gou2024critic}, or use episodic memory buffers to improve decision-making in subsequent attempts \cite{shinn2023reflexion}.

Mixture of Agents (MoA) \cite{wang2025mixtureofagents} is an ensemble approach that combines predictions from multiple different LLMs to improve performance through complementary strengths of diverse models. Self Mixture of Agents (Self-MoA) \cite{li2025rethinking} extends this concept by using multiple instances of the same model with different sampling parameters to generate diverse outputs. While these approaches have shown promise in general language generation tasks, their systematic application to structured prediction tasks like event extraction remains underexplored.

Our work builds upon these advances in LLM fine-tuning and self-correction mechanisms, specifically addressing the challenges of Event Extraction by developing a decomposed instruction fine-tuning approach combined with a structured self-reflection module that enables effective reasoning about event structures. Additionally, we are the first to systematically apply Self-MoA to event extraction, leveraging multiple instances of the same LLM to generate diverse candidate events that are then refined through agreement detection and confidence-based filtering with a discriminative sequence tagger.

\section{Methodology}
\label{methodology}


ARIS aims to enhance event extraction by integrating generative and discriminative approaches through structured model consensus and reflective reasoning. As illustrated in Figure \ref{fig:system_figure}, ARIS initiates with an explicit fine-tuning phase to equip an LLM with specialized capabilities for event subtasks, including trigger identification, event classification, and argument extraction. Following fine-tuning, ARIS implements a Self Mixture of Agents (Self-MoA), leveraging the fine-tuned LLM to generate diverse candidate events. Concurrently, a discriminative sequence tagging model independently predicts events. To consolidate predictions, ARIS employs structured consensus detection and confidence-based filtering, selectively retaining high-confidence agreements while discarding uncertain disagreements. To systematically address remaining ambiguities, ARIS utilizes a reflective inference module that capitalizes on the fine-tuned LLM's contextual reasoning capabilities. The subsequent sections detail the implementation and interactions of these key components. The detailed procedure of our approach is formalized as Algorithm \ref{alg:reflective_agreement}, which can be found in Appendix \ref{sec:app_algorithm}.

\subsection{Self Mixture of Agents for Event Extraction}
\label{sec:decomposed_instruction_tuning}

We formalize our approach with the following notation. Let $A = \{A_1, A_2, \ldots, A_n\}$ be the Self Mixture of Agents, where each agent $A_i$ is an LLM with temperature $T_i$.

\textbf{Event Decomposed Fine-Tuning.} To enhance the LLM's base understanding of the event extraction task, we develop a decomposed instruction fine-tuning dataset to explicitly guide the LLM to master distinct subtasks inherent to event extraction.

Given an event extraction dataset $D_{event} = \{(x_i, y_i)\}_{i=1}^{N}$ (input texts $x_i$ and corresponding event annotations $y_i$), we convert it into a decomposed instructional dataset $D_{decomp}$ structured around three primary subtasks:

First, we create holistic event structure modeling instructions that supervise complete event construction. These include full-structure construction tasks requiring the model to output a complete list of events in the passage (with triggers, types, and arguments with roles), and role-ablated construction variants that systematically mask one argument role per instance, requiring the model to infer the remainder while maintaining structural coherence.

Second, we develop trigger-focused reasoning instructions that isolate the foundational stages of event extraction: trigger detection focuses solely on identifying trigger spans; type classification assigns event types to known triggers (both individually and in batches); trigger discrimination provides binary classification supervision for distinguishing triggers from non-triggers; and joint trigger-type prediction unifies detection and classification into a single structured output.

Third, we create argument-level inference instructions that target post-trigger prediction. Argument extraction requires identifying argument spans for known triggers, role assignment classifies the role of each known argument, and joint argument-role prediction unifies extraction and classification into a single coherent operation.

We fine-tune the initial LLM $M_{LLM}^{init}$ on the decomposed dataset $D_{decomp}$, where the model first learns atomic subtasks (trigger identification, argument extraction) before progressing to intermediate compositional tasks (joint trigger-type prediction, role assignment) and finally to full event structure generation. Further details on this dataset construction can be found in Appendix \ref{appendix:decomp_dataset}.


\subsection{Hybrid Event Aggregation}




Let $S$ be a pretrained sequence tagger (e.g., a fine-tuned transformer such as RoBERTa) trained for event extraction. For an input document $x$ (sentence or article), we define $G_x$ as the space of possible valid event spans in $x$. Each LLM agent $A_i$ produces event extraction predictions $E_{A_i}(x; T_i) \subset G_x$. Simultaneously, the sequence tagger produces its own set of event predictions $E_S(x) \subset G_x$.
Initially, the predictions from all individual LLM agents are aggregated to create a preliminary combined prediction set $E_{SMoA}^{raw}(x) = \bigcup_{i=1}^{n} E_{A_i}(x; T_i)$.







Since the LLM agents generate multiple independent predictions through the self mixture of agents approach, these parallel predictions may identify the same trigger multiple times or reference non-existent spans due to hallucination. To ensure accurate alignment between model predictions and the source text, we apply a rule-based cleanup mechanism that consists of two sequential steps: span validation and positional sorting.

We first filter out predictions that reference non-existent text spans. Let $S_{text}(x) = \{s_1, s_2, \ldots, s_m\}$ be the set of all possible contiguous text spans present in the input document $x$. We retain only predictions whose spans exist in the source text, creating $E_{SMoA}^{valid}(x)$ which contains only events $e$ from $E_{SMoA}^{raw}(x)$ where $\text{span}(e) \in S_{text}(x)$.

We sort the valid predictions by their textual positions to establish a canonical ordering, producing $E_{SMoA}(x) = \text{sort}(E_{SMoA}^{valid}(x), \text{by position in } x)$.

The resulting cleaned prediction set $E_{SMoA}(x)$ serves as the foundation for subsequent agreement detection and disagreement handling steps.





\subsection{Consensus Detection}
We identify consensus predictions as events jointly extracted by both the Self-MoA and sequence tagger, defined as $E_{con}(x) = E_{SMoA}(x) \cap E_S(x)$. Events match when they have the same trigger identification, trigger classification, argument identification and argument classification predictions. These consensus predictions represent the most reliable predictions, where both generative and discriminative methods converge on the same result. Further details on the exact mechanism for consensus detection can be found in Appendix \ref{sec:agreement}.




\subsection{Disagreement Handling via Confidence Filtering}
For cases where the Self-MoA and sequence tagger disagree, we employ a confidence-based filtering strategy. We first define the complete set of predictions $E_{comb}(x) = E_{SMoA}(x) \cup E_S(x)$ and the disagreement set $E_{dis}(x) = E_{comb}(x) \setminus E_{con}(x)$.

We then compute confidence scores for predictions in the disagreement set. To calculate confidence for Self-MoA predictions, we need to track the origin of each prediction. Let $A_e = \{i : e \in E_{A_i}(x; T_i)\}$ be the set of agent indices that predicted event $e$. For Self-MoA predictions, confidence is calculated as the proportion of agents that made the same prediction:

\begin{equation*}
C_{SMoA}(e) = \frac{|A_e|}{n} = \frac{|\{i : e \in E_{A_i}(x; T_i)\}|}{n}
\end{equation*}




For sequence tagger predictions, confidence is derived from the softmax score of the predicted token in the output layer. Let $T_{tags}$ be the set of all possible event tags in the sequence tagger's output vocabulary (including event types and argument roles), and let $\text{span}(e)$ denote the text span associated with event $e$. The confidence is calculated as the maximum probability over all tags for the given span: $C_S(e) = \max_{t \in T_{tags}} P_S(t | \text{span}(e), x)$, where $P_S(t | \text{span}(e), x)$ is the probability distribution over possible tags for the span of event $e$ given the input text $x$.

We define confidence thresholds $\theta_{SMoA}$ and $\theta_S$ to filter out low-confidence predictions. Given the sequence tagger's superior precision in predictions, high-confidence sequence tagger predictions that disagree with the Self-MoA are retained in the final prediction set. Specifically, for sequence tagger predictions with confidence exceeding the threshold ($C_S(e) \geq \theta_S$), we include them directly in the set $E_{hi\_conf}^S(x) = \{e \in E_S(x) \cap E_{dis}(x) \mid C_S(e) \geq \theta_S\}$.

For low-confidence predictions from both models, we apply confidence-based filtering. Let $E_{rem}(x)$ be the set of events to be removed due to low confidence, where $E_{rem}^{SMoA} = \{e \in E_{SMoA}(x) \cap E_{dis}(x) \mid C_{SMoA}(e) < \theta_{SMoA}\}$ represents low-confidence Self-MoA predictions and $E_{rem}^S = \{e \in E_S(x) \cap E_{dis}(x) \mid C_S(e) < \theta_S\}$ represents low-confidence sequence tagger predictions. The complete set of removed events is then $E_{rem}(x) = E_{rem}^{SMoA} \cup E_{rem}^S$.

\subsection{Reflection on Ambiguous Predictions}
The remaining disagreement predictions after confidence filtering represent ambiguous cases that require further analysis. We define this set as $E_{reflect}(x) = E_{dis}(x) \setminus E_{rem}(x)$, capturing all disagreements not removed during filtering.

We resolve these ambiguous cases through a reflection mechanism $R$ that formulates a structured query containing the original text context $x$ and the ambiguous predictions $E_{reflect}(x)$. This query presents each ambiguous prediction along with its surrounding context, asking the LLM to analyze and determine the correct prediction based on linguistic cues, event semantics, and contextual understanding. The reflection process leverages the LLM's reasoning capabilities to produce a refined set of resolved predictions $E_{reflected}(x) = R(E_{reflect}(x), x)$. Further details on this procedure can be found in Appendix \ref{sec:reflection}.

\subsection{Final Prediction Set}
The final prediction set combines high-confidence agreed predictions with those resolved through reflection, forming $E_{fin}(x) = E_{con}(x) \cup E_{hi\_conf}^S(x) \cup E_{reflected}(x)$. This approach leverages the complementary strengths of both generative and discriminative models: the structural consistency and precision of sequence taggers for straightforward cases, and the contextual reasoning capabilities of LLMs for resolving complex ambiguities.

\begin{table*}[t]
\centering
\resizebox{\textwidth}{!}{%
\begin{tabular}{llrrrrrrrrrrrr}
\toprule
\multirow{2}{*}{Base LLM} & \multirow{2}{*}{Approach}
  & \multicolumn{4}{c}{CASIE}
  & \multicolumn{4}{c}{M2E2}
  & \multicolumn{4}{c}{MLEE} \\
\cmidrule(lr){3-6} \cmidrule(lr){7-10} \cmidrule(lr){11-14}
 &  & Trg‐I & Trg‐C & Arg‐I & Arg‐C
      & Trg‐I & Trg‐C & Arg‐I & Arg‐C
      & Trg‐I & Trg‐C & Arg‐I & Arg‐C \\
\midrule
 & TagPrime                             & \textbf{72.00} & \textbf{71.60} & 47.47 & 45.61
                                       & 63.97 & 63.30 & 37.47 & 34.19
                                       & 74.61 & 72.38 & 48.30 & 46.74 \\
\midrule
 & DEBATE-EE                            & –     & 41.80 & –     & 40.50 
                                       & –     & –     & –     & –    
                                       & –     & –     & –     & –     \\
 & MMUTF                                & –     & –     & –     & –    
                                       & –     & 55.50 & –     & 38.20 
                                       & –     & –     & –     & –     \\
\midrule
\multirow{4}{*}{Llama-3.1 8B}
 & One-Shot                             & 0.15  & 0.15  & 0.00  & 0.00  
                                       & 3.77  & 3.77  & 0.40  & 0.40  
                                       & 0.23  & 0.23  & 0.00  & 0.00  \\
 & FineTuned-EE                         & 42.59 & 42.27 & 26.72 & 25.30 
                                       & 66.87 & 63.16 & 31.94 & 27.78 
                                       & 50.64 & 44.96 & 38.75 & 34.93 \\
 & FineTuned-DEE                        & 65.89 & 65.34 & 44.60 & 42.56 
                                       & 67.43 & 64.00 & 36.33 & 32.80 
                                       & 55.40 & 52.14 & 50.98 & 48.33 \\
 & ARIS                                 & 70.78 & 70.27 & \textbf{48.79} & \textbf{46.84}
                                       & 73.49 & \textbf{71.39} & \textbf{41.41} & \textbf{38.84}
                                       & 73.80 & 70.33 & 54.30 & 50.86 \\
\midrule
\multirow{4}{*}{Phi-3 7B}
 & One-Shot                             & 3.11  & 2.49  & 0.85  & 0.60  
                                       & 31.37 & 27.45 & 9.35  & 4.68  
                                       & 1.13  & 0.68  & 0.72  & 0.72  \\
 & FineTuned-EE                         & 41.99 & 41.32 & 26.83 & 25.74 
                                       & 59.44 & 55.11 & 28.14 & 24.31 
                                       & 29.11 & 26.53 & 26.98 & 24.25 \\
 & FineTuned-DEE                        & 62.26 & 61.53 & 42.00 & 40.55 
                                       & 67.26 & 64.31 & 32.30 & 29.57 
                                       & 37.89 & 35.52 & 45.93 & 44.31 \\
 & ARIS                                 & 69.08 & 68.39 & 46.63 & 44.99
                                       & \textbf{74.22} & \textbf{71.39} & 41.11 & 36.61
                                       & \textbf{74.78} & \textbf{72.45} & \textbf{59.19} & \textbf{56.98} \\
\bottomrule
\end{tabular}%
}
\caption{F1 score of Event Extraction performance (Trg=trigger, Arg=argument; I=identification, C=classification) across three benchmark datasets. Bold numbers indicate best performance on evaluation metric.}
\label{tab:main_experiment}
\end{table*}

As shown in Figure \ref{fig:system_figure}, our framework effectively handles hallucinations and disagreements between models. For instance, in the example text "The prosecutor, Alberto Nisman, was found shot dead in his bathroom in January - four days after he accused Fernandez and her aides of making a deal with Iran to cover up the alleged roles that Iranian officials played in the 1994 bombing of a Jewish center in Argentina.", the sequence tagger identifies only the trigger ['bombing'], while the MoA detects multiple candidates including ['dead', 'shot', 'bombing']. Through our reflective agreement process, the incorrect trigger 'shot' is filtered out, resulting in the accurate final triggers ['dead', 'bombing']. This demonstrates how ARIS combines discriminative precision with generative coverage while eliminating hallucinations.

\section{Experiments}
\label{experiments}

\subsection{Dataset and Evaluation Metrics}

We evaluate our approach on three benchmark datasets for event extraction processed following the TextEE benchmark standardization process \cite{huang2024textee}: CASIE \cite{satyapanich2020casie}, M2E2 \cite{li-etal-2020-cross}, and MLEE \cite{pyysalo2012event}. We use the train/dev/test partitions defined in TextEE's "split1" for all three datasets. For M2E2, we used only the text, and did not include any image or video information. These datasets represent diverse domains and text structures: CASIE covers cybersecurity news with 5 event types in long paragraphs; M2E2 contains shorter news content with 8 event types primarily in 1-2 sentence format; and MLEE represents the biomedical domain with 29 event types across long paragraphs. 

For evaluation, we report micro F1 scores for the following tasks: Trigger Identification, which evaluates the model's ability to correctly identify event trigger spans in text, regardless of event type; Trigger Classification, which measures performance in both identifying event triggers and correctly classifying their event types; Argument Identification, which assesses the model's capability in identifying argument entities associated with correctly identified event triggers; and Argument Classification, which requires correct identification of both the argument entity and its role assignment for a given event trigger. For all metrics, we employ exact match scoring.

\subsection{Implementation}

For the discriminative sequence tagger component of our proposed approach, we utilize TagPrime \cite{hsu2023tagprime}, a unified framework for relational structure extraction that has demonstrated superior performance on event extraction tasks. We implement TagPrime with roberta-large from huggingface \cite{roberta_large_hf_2024} as the backbone encoder \cite{liu2019roberta}.

For the generative component, we employ Llama-3.1-8B-Instruct \cite{grattafiori2024llama} and Phi-3-small-8k-instruct \cite{abdin2024phi}. We access both models through their respective Hugging Face implementations \cite{llama_31_hf, phi_3_hf}. To train both LLMs for event extraction, we used LoRA \cite{hu2022lora} implemented with the Hugging Face PEFT library \cite{peft}. Our LoRA configuration uses a rank of 32, a scaling factor ($\alpha$) of 128, and a dropout rate of 0.05. In all of our experiments we utilize 10 Self-MoA Agents that have a temperature of 0.9 unless stated otherwise. For our confidence-based filtering for our ARIS approach, we dynamically determined dataset-specific thresholds for each model and temperature setting to optimize system performance. These thresholds and details on how they were computed, are reported in Appendix~\ref{appendix:confidence_thresholds}.

\begin{table*}[t]
\centering
\resizebox{\textwidth}{!}{%
\begin{tabular}{lllrrrrrrrrrrrrr}
\toprule
\multirow{2}{*}{Base LLM} & \multirow{2}{*}{Approach} & \multirow{2}{*}{Temp.}
  & \multicolumn{3}{c}{Trg‐I}
  & \multicolumn{3}{c}{Trg‐C}
  & \multicolumn{3}{c}{Arg‐I}
  & \multicolumn{3}{c}{Arg‐C} \\
\cmidrule(lr){4-6}\cmidrule(lr){7-9}\cmidrule(lr){10-12}\cmidrule(lr){13-15}
 &  &  & P & R & F1 & P & R & F1 & P & R & F1 & P & R & F1 \\
\midrule
\multirow{6}{*}{Llama‐3.1 8B}
 & Self‐MoA & 0.9 & 46.99 & 71.17 & 56.51 & 44.98 & 68.29 & 54.15 & 28.03 & 63.15 & 38.76 & 26.28 & 59.09 & 36.32 \\
 & Self‐MoA & 0.6 & 53.12 & 68.42 & 59.72 & 50.99 & 65.79 & 57.37 & 30.73 & 58.79 & 40.33 & 28.81 & 55.02 & 37.80 \\
 & Self‐MoA & 0.1 & 63.18 & 61.31 & 62.07 & 60.79 & 59.02 & 59.75 & 39.25 & 52.83 & 44.98 & 37.19 & 50.03 & 42.61 \\
 & ARIS      & 0.9 & 69.16 & 76.85 & 72.69 & 67.20 & 74.75 & 70.66 & 42.94 & 55.27 & 48.17 & 40.60 & 52.19 & 45.51 \\
 & ARIS      & 0.6 & 68.26 & 76.59 & 72.15 & 66.37 & 74.52 & 70.17 & 48.56 & 50.52 & 49.37 & 46.30 & 48.04 & 46.94 \\
 & ARIS      & 0.1 & 66.72 & 77.19 & 71.56 & 64.72 & 74.92 & 69.43 & 44.78 & 53.06 & 48.38 & 42.61 & 50.53 & 46.59 \\
\midrule
\multirow{6}{*}{Phi‐3 7B}
 & Self‐MoA & 0.9 & 39.02 & 73.50 & 50.47 & 37.15 & 70.34 & 48.13 & 24.80 & 60.23 & 35.08 & 23.30 & 56.55 & 32.95 \\
 & Self‐MoA & 0.6 & 46.29 & 69.46 & 55.02 & 44.27 & 66.75 & 52.72 & 32.35 & 58.62 & 41.54 & 30.69 & 55.43 & 39.37 \\
 & Self‐MoA & 0.1 & 59.77 & 60.80 & 59.82 & 57.15 & 58.35 & 57.30 & 41.40 & 49.14 & 44.62 & 39.52 & 46.73 & 42.52 \\
 & ARIS      & 0.9 & 71.55 & 74.81 & 72.69 & 69.58 & 72.86 & 70.74 & 47.16 & 51.65 & 48.79 & 44.52 & 48.64 & 46.00 \\
 & ARIS      & 0.6 & 73.49 & 74.52 & 73.78 & 71.46 & 72.54 & 71.78 & 50.01 & 49.33 & 49.43 & 47.68 & 46.94 & 47.06 \\
 & ARIS      & 0.1 & 68.84 & 76.03 & 72.04 & 66.70 & 73.77 & 69.85 & 46.64 & 50.76 & 48.38 & 44.63 & 48.45 & 46.24 \\
\bottomrule
\end{tabular}%
}
\caption{Impact of sampling temperature on event extraction performance. Results show precision (P), recall (R), and F1 scores across different sampling temperatures.}
\label{tab:aris-temp}
\end{table*}

\subsection{Overall Event Extraction Performance}

We compare our proposed event extraction approach against several baseline approaches and state-of-the-art methods across multiple configurations. Our baselines include various LLM-based approaches using both zero-shot and few-shot prompting, as well as fine-tuned variants. The One-Shot baseline employs off-the-shelf LLMs with carefully designed prompts that explain the event extraction task and dataset structure, and provide a single event extraction example without any task-specific training. For each test instance, we selected the most similar training example using TF-IDF vectorization \cite{salton1988term} with cosine similarity, ensuring that the provided examples are contextually relevant to the test cases.

For fine-tuned approaches, we implement two training strategies: FineTuned-EE represents standard end-to-end fine-tuning where the LLM learns to directly map input text to complete event structures (event triggers, event types, event arguments, and event argument roles) in a single step. In contrast, FineTuned-DEE leverages our proposed decomposed instruction fine-tuning approach (Section \ref{sec:decomposed_instruction_tuning}), where the model first learns individual subtasks before progressing to complete event extraction.  For both FineTuned-EE and FineTuned-DEE, inference is performed using a single LLM with a temperature setting of 0.9. Our final proposed method, ARIS, combines the decomposed fine-tuning with our proposed inference framework that integrates the Self Mixture of Agents, consensus detection, confidence-based filtering, and reflection mechanisms.

We compare against several strong baselines including TagPrime \cite{hsu2023tagprime}, a discriminative sequence tagging model that represents the current state-of-the-art for event extraction tasks, as well as recent LLM-based approaches: DEBATE-EE \cite{wang-huang-2024-debate}, which employs a multi-agent debate framework that iteratively refines event extraction predictions through discussions between debating agents, critics, and judges, enhanced with diverse retrieval-augmented generation and adaptive conformal prediction modules, and MMUTF \cite{seeberger-etal-2024-mmutf}, a unified template filling framework that extracts event arguments by matching candidates to argument roles using templates as queries. The results shown for DEBATE-EE and MMUTF rows are the F1 scores provided in their original papers.

The results in Table \ref{tab:main_experiment} demonstrate that our proposed approach achieves significant improvements over baseline methods across all three datasets. Our ARIS approach consistently outperforms competing methods, particularly in argument extraction tasks.

On CASIE, our method is competitive with TagPrime for trigger detection while surpassing it on argument tasks. For M2E2, we achieve the strongest performance across all metrics, with both Llama-3.1 and Phi-3 implementations of ARIS significantly outperforming TagPrime. On MLEE, our Phi-3 based ARIS implementation shows the strongest performance across all metrics. Notably, ARIS with Phi-3 improves argument classification F1 scores over the TagPrime model, surpassing the strong baseline by over 10 points.

The results of the one-shot experiment show that LLMs may struggle with event extraction when using basic prompting strategies, showing the need for fine-tuning on the task. The effectiveness of decomposed instruction fine-tuning is evident when comparing FineTuned-DEE with standard FineTuned-EE, showing consistent improvements across all datasets. The ARIS framework further enhances performance, particularly for argument-related tasks. These results validate our hypothesis that combining the complementary strengths of discriminative models and LLMs through our reflective agreement approach effectively addresses the limitations of individual approaches.

\begin{table*}[t]
\centering
\resizebox{\textwidth}{!}{%
\begin{tabular}{llrrrrrrrrrrrr}
\toprule
\multirow{2}{*}{Base LLM} & \multirow{2}{*}{Approach}
  & \multicolumn{3}{c}{Trg‐I}
  & \multicolumn{3}{c}{Trg‐C}
  & \multicolumn{3}{c}{Arg‐I}
  & \multicolumn{3}{c}{Arg‐C} \\
\cmidrule(lr){3-5}\cmidrule(lr){6-8}\cmidrule(lr){9-11}\cmidrule(lr){12-14}
 &  & P & R & F1 & P & R & F1 & P & R & F1 & P & R & F1 \\
\midrule
\multirow{3}{*}{Llama‐3.1 8B}
 & Self-MoA & 46.99 & 71.17 & 56.51 & 44.98 & 68.29 & 54.15 & 28.03 & \textbf{63.15} & 38.76 & 26.28 & \textbf{59.09} & 36.32 \\
 & ARIS w/o TagPrime              & 60.87 & 60.08 & 60.31 & 58.69 & 58.08 & 58.23 & 33.26 & 57.10 & 42.03 & 31.38 & 53.87 & 39.66 \\
 & ARIS                           & 69.16 & \textbf{76.85} & \textbf{72.69} & 67.20 & \textbf{74.75} & 70.66 & 42.94 & 55.27 & 48.17 & 40.60 & 52.19 & 45.51 \\
\midrule
\multirow{3}{*}{Phi-3 7B}
 & Self-MoA & 39.02 & 73.50 & 50.47 & 37.15 & 70.34 & 48.13 & 24.80 & 60.23 & 35.08 & 23.30 & 56.55 & 32.95 \\
 & ARIS w/o TagPrime              & 53.52 & 64.50 & 57.52 & 50.84 & 61.65 & 54.78 & 31.30 & 52.44 & 38.53 & 29.38 & 49.55 & 36.26 \\
 & ARIS                           & 71.55 & 74.81 & \textbf{72.69} & 69.58 & 72.86 & \textbf{70.74} & 47.16 & 51.65 & \textbf{48.79} & 44.52 & 48.64 & \textbf{46.00} \\
\midrule
 & TagPrime                       & \textbf{76.64} & 65.33 & 70.19 & \textbf{75.43} & 64.32 & 69.09 & \textbf{49.34} & 40.88 & 44.41 & \textbf{46.76} & 38.88 & 42.18 \\
\bottomrule
\end{tabular}%
}
\caption{Ablation study demonstrating component contributions to event extraction performance. Results highlight the complementary strengths of discriminative (TagPrime) and generative approaches.}
\label{tab:ablation}
\end{table*}

\subsection{Impact of Temperature on Self-MoA LLMs}

To understand how sampling diversity impacts performance, we evaluated our approach across three temperature settings (t=0.9, t=0.6, and t=0.1) during inference. Table \ref{tab:aris-temp} presents the averaged results across all datasets.

The results reveal that while temperature substantially affects standalone Self-MoA performance, the full ARIS approach maintains consistent performance across all settings. For Self-MoA alone, temperature variations produce dramatic differences in precision-recall trade-offs, where higher temperatures lead to higher recall and lower precision. We observe dramatic shifts in precision-recall balance for Phi-3, with precision increasing from 39.02\% to 59.77\% for trigger identification while recall drops from 73.50\% to 60.80\%.

In contrast, our full ARIS approach demonstrates stability across the tested temperatures, with F1 scores for all metrics showing variation of less than 2 points. This stability demonstrates that our confidence-based filtering, agreement detection, and reflection mechanisms effectively normalize the varying predictions produced at different temperatures.

\subsubsection{Ablation Study}

To understand the contributions of different components in our approach, we conducted an ablation study focusing on the integration of the sequence tagger with the ARIS framework. Table \ref{tab:ablation} presents results averaged across all datasets, comparing our full ARIS approach against variants with components removed.

The results reveal clear complementary strengths between the discriminative and generative components. The RoBERTa-based TagPrime sequence tagger demonstrates superior precision across all tasks, but shows lower recall. Conversely, the ARIS approach without TagPrime or reflection exhibits higher recall, but show lower precision.

Our full ARIS approach effectively leverages these complementary strengths that results in higher F1 scores across all metrics. The improvements are even more pronounced for argument-related tasks. The practical impact of these complementary strengths is demonstrated through detailed pipeline examples in Appendix \ref{sec:examples}. These examples trace the complete processing flow from initial predictions through agreement detection, confidence filtering, and reflection, providing concrete illustrations of how ARIS effectively leverages the precision of discriminative models and the semantic flexibility of generative approaches to improve event extraction performance.


\section{Conclusion}
\label{conclusion}
In this paper, we introduced ARIS, a hybrid event extraction method that combines the complementary strengths of discriminative sequence taggers and generative LLMs through a structured reflective agreement mechanism. Our approach leverages a Self Mixture of Agents to generate diverse event predictions, employs agreement detection to identify high-confidence consensus predictions, applies confidence-based filtering to eliminate low-precision candidates, and utilizes a reflection mechanism powered by decomposed instruction fine-tuning to resolve ambiguous cases. Experiments across three benchmark datasets demonstrate that ARIS consistently outperforms existing state-of-the-art methods, with particularly notable improvements in argument extraction tasks. Beyond empirical performance gains, our work advances the theoretical understanding of hybrid model integration and structured reflective reasoning in complex NLP tasks.

\section*{Limitations}
\label{limitations}

While our proposed approach demonstrates improvements in event extraction performance, it comes with significant computational overhead. The approach requires running multiple LLM instances for the Self-MoA component, which substantially increases both inference time and computational resources compared to traditional discriminative models. Additionally, the training process for decomposed instruction fine-tuning demands considerable GPU resources and time, particularly when working with larger LLMs. A further limitation concerns the reflection mechanism itself, which handles the most ambiguous cases that neither the Self-MoA nor discriminative model could confidently resolve. While effective for some difficult instances, the reflection component may still struggle with highly challenging cases. This is an inherent ceiling to the reflection-based approach when confronted with the hardest examples.

\bibliography{anthology,custom}

\appendix

\section{Agreement Detection}
\label{sec:agreement}

\noindent Agreement detection reconciles event predictions from the Self-MoA ensemble and the sequence tagger by identifying cases where both systems refer to the same underlying event mention. This process involves separate matching criteria for triggers and arguments, as detailed below.

\subsection{Trigger Span Agreement} 
Trigger predictions from both models are considered to be in agreement if their textual spans overlap beyond a predefined threshold, indicating they refer to the same underlying event mention. This criterion effectively handles partial overlaps, such as when one span is a substring of another—for example, "\texttt{attached}" versus "\texttt{was attached}"—as they semantically represent the same trigger. In these scenarios, the span predicted by the sequence tagger is retained due to its higher precision in determining exact span boundaries.

\subsection{Argument Span Agreement} 
To determine agreement between argument predictions, we first align them based on their associated trigger and event type. For each matched trigger between the two systems, its candidate arguments are evaluated independently for span-level overlap. This enables partial agreement at the argument level: a single trigger may have some arguments in agreement and others not, depending on their span overlap. For example, arguments like "\texttt{the government officials}" and "\texttt{government officials}" linked to the same trigger and event type are considered to be in agreement. In such cases, we retain the span predicted by the sequence tagger due to its higher precision in boundary identification.

\section{Confidence Threshold Selection}
\label{appendix:confidence_thresholds}
\setlength{\parindent}{0pt}

Predictions that are not in agreement between the two systems enter this phase. Confidence thresholds for filtering event predictions were dataset-specific, determined by analyzing the distribution of confidence scores on validation sets. For each dataset, we:

\begin{enumerate}
\setlength{\itemsep}{0.5ex}   
\setlength{\parskip}{0pt}     
\setlength{\topsep}{1ex}
\item Computed confidence score distributions separately for correct (found in gold annotations) and incorrect predictions.
\item Used descriptive statistics (mean, median, quartiles) to guide a targeted search range for optimal thresholds.
\item Conducted a search within this range, selecting thresholds that maximized the validation set F$_1$ score.
\end{enumerate}

This threshold selection procedure was repeated individually for trigger and argument predictions, ensuring dataset-specific tuning that improved overall performance. After the agreement detection phase, predictions identified as disagreements are handled based on finalized confidence thresholds: high-confidence disagreements are retained directly, low-confidence disagreements are discarded immediately, and intermediate-confidence cases are forwarded to the reflection mechanism for further analysis.

\subsection{Dataset and Temperature Specific Confidence Thresholds}
\label{appendix:confidence_tables}

These two tables present the per-dataset, per-temperature confidence thresholds applied during disagreement handling.  Table~\ref{tab:trigger_thresholds} gives the trigger-level thresholds, while Table~\ref{tab:argument_thresholds} lists the argument-level thresholds.  In both tables, $\theta_{S}$ is the sequence-tagger (TagPrime) retention threshold, $\theta^{+}_{\text{SMoA}}$ is the high-confidence Self-MoA keep threshold, and $\theta^{-}_{\text{SMoA}}$ is the low-confidence Self-MoA drop threshold.

\begin{table}[h]
  \centering\small
  \begin{tabular}{llcccc}
    \toprule
    \textbf{Model} & \textbf{Dataset} & \textbf{Temp} & $\theta_{S}$ & $\theta^{+}_{\text{SMoA}}$ & $\theta^{-}_{\text{SMoA}}$ \\
    \midrule
    \multirow{9}{*}{Phi-3} 
      & M2E2 & 0.1 & 0.90 & 0.90 & 0.20 \\
      &      & 0.6 & 0.89 & 0.95 & 0.60 \\
      &      & 0.9 & 0.89 & 0.90 & 0.50 \\ 
    \cmidrule(lr){2-6}
      & CASIE & 0.1 & 0.035 & 1.10 & 0.90 \\
      &       & 0.6 & 0.008 & 1.10 & 0.80 \\
      &       & 0.9 & 0.007 & 0.95 & 0.40 \\
    \cmidrule(lr){2-6}
      & MLEE & 0.1 & 0.99 & 1.10 & 0.90 \\
      &      & 0.6 & 0.99 & 1.10 & 0.90 \\
      &      & 0.9 & 0.99 & 1.10 & 0.80 \\
    \midrule
    \multirow{9}{*}{Llama-3.1} 
      & M2E2 & 0.1 & 0.80 & 1.00 & 0.70 \\
      &      & 0.6 & 0.80 & 0.90 & 0.50 \\
      &      & 0.9 & 0.80 & 0.85 & 0.30 \\ 
    \cmidrule(lr){2-6}
      & CASIE & 0.1 & 0.004 & 1.00 & 0.90 \\
      &       & 0.6 & 0.004 & 0.95 & 0.50 \\
      &       & 0.9 & 0.004 & 0.90 & 0.50 \\
    \cmidrule(lr){2-6}
      & MLEE & 0.1 & 1.00 & 1.00 & 0.10 \\
      &      & 0.6 & 1.00 & 0.95 & 0.40 \\
      &      & 0.9 & 0.00 & 0.80 & 0.55 \\
    \bottomrule
  \end{tabular}
  \caption{Trigger-level confidence thresholds}
  \label{tab:trigger_thresholds}
\end{table}

\begin{table}[h]
  \centering\small
  \begin{tabular}{llcccc}
    \toprule
    \textbf{Model} & \textbf{Dataset} & \textbf{Temp} & $\theta_{S}$ & $\theta^{+}_{\text{SMoA}}$ & $\theta^{-}_{\text{SMoA}}$ \\
    \midrule
    \multirow{9}{*}{Phi-3} 
      & M2E2 & 0.1 & 1.00 & 1.00 & 0.70 \\
      &      & 0.6 & 0.99 & 0.85 & 0.60 \\
      &      & 0.9 & 0.99 & 0.75 & 0.30 \\ 
    \cmidrule(lr){2-6}
      & CASIE & 0.1 & 0.07 & 1.10 & 0.95 \\
      &       & 0.6 & 1.10 & 0.95 & 0.50 \\
      &       & 0.9 & 0.07 & 0.81 & 0.30 \\
    \cmidrule(lr){2-6}
      & MLEE & 0.1 & 1.10 & 0.90 & 0.30 \\
      &      & 0.6 & 1.10 & 0.90 & 0.40 \\
      &      & 0.9 & 1.10 & 0.60 & 0.30 \\
    \midrule
    \multirow{9}{*}{Llama-3.1} 
      & M2E2 & 0.1 & 0.99 & 0.99 & 0.50 \\
      &      & 0.6 & 0.99 & 1.00 & 0.90 \\
      &      & 0.9 & 0.90 & 0.99 & 0.50 \\ 
    \cmidrule(lr){2-6}
      & CASIE & 0.1 & 0.05 & 1.00 & 0.70 \\
      &       & 0.6 & 0.03 & 0.90 & 0.60 \\
      &       & 0.9 & 0.04 & 0.90 & 0.60 \\
    \cmidrule(lr){2-6}
      & MLEE & 0.1 & 1.00 & 0.70 & 0.50 \\
      &      & 0.6 & 1.00 & 0.80 & 0.50 \\
      &      & 0.9 & 1.00 & 0.50 & 0.10 \\
    \bottomrule
  \end{tabular}
  \caption{Argument-level confidence thresholds}
  \label{tab:argument_thresholds}
\end{table}

\section{Reflection Mechanism}
\label{sec:reflection}
Our reflection mechanism addresses ambiguous predictions, cases of disagreement between the Self-MoA ensemble and the sequence tagger that remain unresolved after confidence-based filtering. This section details the structured reflection procedure, including prompt design, parsing strategies, and LLM configuration.

\subsection{Prompt Format}
\label{appendix:structured_query}
Our reflection mechanism employs carefully designed structured prompts to elicit precise responses from the LLM. Each prompt follows a format comprising:
\begin{enumerate}
\setlength{\itemsep}{0.5ex}   
\setlength{\parskip}{0pt}     
\setlength{\topsep}{1ex}
\item \textbf{Role specification}: Defines the LLM's precise role (e.g., argument validator).
\item \textbf{Task description}: Provides explicit instructions for classifying candidates.
\item \textbf{Generation rules}: Sets strict output constraints to avoid hallucinations and ensure structured responses.
\item \textbf{Context}: Supplies the complete passage and candidate triggers or arguments for accurate contextual evaluation.
\item \textbf{Example output}: Demonstrates the required structured output format.
\end{enumerate}

\subsection{LLM Configuration}
\label{appendix:llm_config}
For both trigger and argument reflection, we use our fine-tuned LLMs with the following settings to ensure deterministic and accurate outputs: \begin{itemize}
\setlength{\itemsep}{0.5ex}   
\setlength{\parskip}{0pt}     
\setlength{\topsep}{1ex}
\item \textbf{Temperature}: 0.1 (to ensure consistent, deterministic outputs)
\item \textbf{Max tokens}: 4096
\item \textbf{Length penalty}: 1.05 (to maintain concise, focused responses)
\end{itemize}

\subsection{Trigger Reflection}
Triggers requiring reflection are presented within structured prompts (see Figure~\ref{fig:reflection-prompt}). The LLM classifies each candidate as "\texttt{Trigger}" or "\texttt{Non-Trigger}". Parsed reflection results update the final trigger set by retaining only confirmed triggers for subsequent argument extraction.

\begin{figure}[t]
  \centering
  \begin{lstlisting}[xleftmargin=0pt,breakindent=0pt,breakautoindent=false]
You previously identified the following candidate triggers:

<CANDIDATE_TRIGGERS_TO_VERIFY>

Your task is to decide for each whether it truly signals an event trigger.

Generation Rules:
1. Classify each phrase as either 'Trigger' or 'Non-Trigger'.
2. Output strictly in the required format-no extra text.

Output Format (strict):
- Wrap the answer in triple backticks (```)
- Write: ClassificationMap = {"phrase1": "Trigger", "phrase2": "Non-Trigger", ...}

Example:
```ClassificationMap = {"therapy": "Trigger", "increase dose": "Non-Trigger"}```

Passage:
<FULL_PASSAGE_TEXT>

Candidates:
<TRIGGER_CANDIDATE_LIST>

Q: For each candidate above, decide whether it is a 'Trigger' or 'Non-Trigger'.
  \end{lstlisting}
  \caption{Structured prompt for binary trigger verification via reflection.}
  \label{fig:reflection-prompt}
\end{figure}

\subsection{Argument Reflection}
Ambiguous arguments undergo similar reflection prompts (Figure~\ref{fig:argument-reflection-prompt}), explicitly linking each argument to its trigger. The LLM assigns a binary \texttt{is\_correct} flag, enabling precise filtering and integration into final event representations.

\begin{figure}[t]
  \centering
  \begin{lstlisting}[xleftmargin=0pt,breaklines=true,breakindent=0pt,breakautoindent=false]
You are an argument validator.
Given a single trigger and its candidate arguments, decide which arguments are valid.

Generation Rules:
1. An argument is valid only if the passage supports its role for this trigger.
2. Preserve the input order-do not add, remove, or reorder.
3. Output exactly three fields per argument: `text`, `role`, `is_correct`.
4. Wrap the entire response in triple backticks (```).

Passage:
"<FULL_PASSAGE_TEXT>"

Trigger:
"<TRIGGER_TEXT>" (type: "<EVENT_TYPE>")

Candidate Arguments to verify:
<CANDIDATE_ARGUMENTS_TO_VERIFY>

Q: For each candidate above, set `is_correct` to `true` or `false`.
  \end{lstlisting}
  \caption{Structured prompt for binary argument verification via reflection.}
  \label{fig:argument-reflection-prompt}
\end{figure}



\section{Final Integration of Predictions}
\label{appendix:integration}
\setlength{\parindent}{0pt}

After performing agreement detection, confidence-based filtering, and reflection, we consolidate predictions into a unified output representation.

\subsection{Triggers}
Trigger predictions fall into one of three categories:
\begin{itemize}
    \setlength{\itemsep}{0.5ex}   
    \setlength{\parskip}{0pt}     
    \setlength{\topsep}{1ex}
    \item \textbf{Agreed triggers:} Identified by both the Self-MoA ensemble and the sequence tagger.
    \item \textbf{High-confidence single-source triggers:} Produced by only one model but retained due to exceeding the confidence threshold.
    \item \textbf{Reflected triggers:} Ambiguous cases resolved by the LLM reflection mechanism.
\end{itemize}

Each group is maintained as a separate list during processing. In the final stage, all triggers are merged to form the complete trigger set for each document.

\subsection{Arguments}
Arguments are integrated per trigger, preserving the provenance of each prediction. For a given trigger, its associated arguments may come from any of the following sources:
\begin{itemize}
    \setlength{\itemsep}{0.5ex}   
    \setlength{\parskip}{0pt}     
    \setlength{\topsep}{1ex}
    \item \textbf{Agreed arguments:} Confirmed by both systems for a shared trigger.
    \item \textbf{High-confidence disagreements:} Provided by one system with sufficient confidence.
    \item \textbf{Reflected arguments:} Verified after reflection over ambiguous trigger-argument pairs.
\end{itemize}

Throughout processing, argument predictions carry the identifier of their associated trigger, allowing us to correctly reassemble arguments under their originating triggers during the final merge. This ensures that each trigger in the final output is paired with the full set of validated and reconciled arguments, regardless of their source path in the pipeline.

\section{ARIS Algorithm}
\label{sec:app_algorithm}

The Reflective Agreement algorithm integrates predictions from a discriminative model (TagPrime) and a generative Mixture-of-Agents (Self-MoA), systematically leveraging model consensus, confidence-based filtering, and reflective inference to enhance event extraction accuracy. The ARIS Algorithm can be found in Algorithm \ref{alg:reflective_agreement}.

\begin{algorithm*}[ht]
\caption{Reflective Agreement for Event Extraction (ARIS)}
\label{alg:reflective_agreement}
\DontPrintSemicolon
\KwData{Ordered trigger sets $E_{tagger}$ and $E_{SMoA}$, confidence threshold $\tau$, Reflection Module $R$, input text $x$}
\KwResult{Final event trigger set $E_{final}(x)$}

$E_{agree}(x) \gets \{(t,p) \mid (t,p,\_) \in E_{tagger}, (t,p,\_) \in E_{SMoA}\}$

$E_{disagree}(x) \gets \{(t,p) \mid (t,p,\_) \in (E_{tagger} \cup E_{SMoA}) \setminus E_{agree}(x)\}$

$E_{high\_conf}(x), E_{ambiguous}(x) \gets \emptyset, \emptyset$

\ForEach{$(t,p) \in E_{disagree}(x)$}{
    $conf_{tagger} \gets$ confidence of $(t,p)$ in $E_{tagger}$ ($0$ if missing)

    $conf_{SMoA} \gets$ confidence of $(t,p)$ in $E_{SMoA}$ ($0$ if missing)

    $combined\_conf \gets \frac{conf_{tagger} + conf_{SMoA}}{\text{number of models predicting }(t,p)}$

    \If{$combined\_conf \geq \tau$}{
        $E_{high\_conf}(x) \gets E_{high\_conf}(x) \cup \{(t,p)\}$
    }
    \Else{
        $E_{ambiguous}(x) \gets E_{ambiguous}(x) \cup \{(t,p)\}$
    }
}

$E_{reflected}(x) \gets R(E_{ambiguous}(x), x)$

$E_{final}(x) \gets \text{SortByPosition}(E_{agree}(x) \cup E_{high\_conf}(x) \cup E_{reflected}(x))$

\Return{$E_{final}(x)$}
\end{algorithm*}

\section{Decomposed Instruction Dataset Construction}
\label{appendix:decomp_dataset}


This appendix describes the construction of the Decomposed Instruction Dataset that is used for the instruction fine‐tuning stage in ARIS (Section~\ref{sec:decomposed_instruction_tuning}).  The goal of this dataset is to teach the LLM the complete reasoning chain of event extraction through a curriculum of thirteen task variants. To equip the LLMs with a rich understanding of each event extraction subtask. We curated instruction datasets from MLEE, M2E2, and CASIE, converting each into a unified JSON schema following the TextEE split 1 configuration.

\paragraph{Holistic Event-Structure Modeling}
These variants require the model to generate an end-to-end representation of every event in a passage, enforcing coherence across triggers, types, and arguments:
\begin{itemize}
    \setlength{\itemsep}{0.5ex}   
    \setlength{\parskip}{0pt}     
    \setlength{\topsep}{1ex}
  \item \textbf{Full‐Structure Construction}: extract all triggers in passage order, assign each the correct event type, and list every argument with its role.
  \item \textbf{Role‐Ablated Construction}: as above, but systematically mask exactly one argument role per instance, compelling the model to infer missing components.
\end{itemize}

\paragraph{Trigger-Focused Reasoning}
By isolating the foundational stage of event extraction, these variants sharpen the model’s precision in identifying and classifying triggers:
\begin{itemize}
    \setlength{\itemsep}{0.5ex}   
    \setlength{\parskip}{0pt}     
    \setlength{\topsep}{1ex}
  \item \textbf{Trigger Detection Only}: list every trigger span in passage order, without type information.
  \item \textbf{Trigger Type Classification – Single}: given one trigger, choose its event type.
  \item \textbf{Trigger Type Classification – Multi}: batch‐classify the types of all triggers.
  \item \textbf{Trigger vs.\ Non-Trigger Discrimination}: binary classification of candidate n-grams as triggers or non-triggers, using hard negatives drawn from the local context.
    \item \textbf{Event Detection (joint)}: detect all triggers and assign types within a single structured output.

\end{itemize}

\paragraph{Argument-Level Inference}
Focusing on post-trigger reasoning, these variants train the model to extract and label arguments conditioned on known triggers:
\begin{itemize}
    \setlength{\itemsep}{0.5ex}   
    \setlength{\parskip}{0pt}     
    \setlength{\topsep}{1ex}
  \item \textbf{Argument Extraction – Single}: list all argument spans for one specified trigger (roles omitted).
  \item \textbf{Argument Extraction – Multi}: for each trigger in passage order, list its arguments (roles omitted).
  \item \textbf{Role Assignment – Single}: given one trigger–argument pair, assign the correct role.
  \item \textbf{Role Assignment – Multi}: for a specified trigger, assign roles to all its candidate arguments in order.
    \item \textbf{Argument Extraction (Joint)}: Given all triggers, extract all associated arguments for each trigger and assign a semantic role to each.
\end{itemize}

Table~\ref{tab:decomp-dataset-stats} summarizes the number of instruction examples per variant and dataset, illustrating the scale and balance of our decomposed curriculum. Together, these thirteen variants provide a curriculum that progresses from atomic subtasks (e.g., isolated classification) to full event construction.

All prompts follow a six-part canonical structure (role → task → rules → format → example → query) and answers are serialized as fenced code blocks under a single top-level key (e.g., \texttt{EventArguments}, \texttt{Triggers}, \texttt{RoleAssignments}).

\subsection{Negative Sampling for Trigger Discrimination}
To generate \texttt{trigger vs.\ non-trigger} examples, we sample negative n-grams that (i) occur exactly once in the passage, (ii) share no substring with any gold trigger, (iii) lie within a three-token window of any trigger, and (iv) satisfy a POS constraint (verbs, nouns, or determiners).  We draw up to three negatives per document to ensure sufficient coverage of hard negatives.

\subsection{Example Instruction}
Figure~\ref{fig:arg-extraction-single-example} presents a complete \emph{Argument Extraction – Single} instruction.

\begin{figure}[p]
    \centering
    \begin{lstlisting}[xleftmargin=0pt,breakindent=0pt,breakautoindent=false]
You are an argument extractor.
Extract all arguments for the specific trigger shown below.

Generation Rules:
1. List arguments in the exact order they appear in the passage.
2. Ignore argument roles and include only the argument texts.

Output Format (strict):
- Wrap the answer in triple backticks (```).
- Write: Arguments = ["arg1", "arg2", ...].

Example:
```
Arguments = ["insulin", "VEGF"]
```

Passage:
"US Needs Broad Coalition to Fight IS Militants, Analysts Say-With President Barack Obama setting a new strategy to combat Islamic State militants (also known as ISIL or ISIS) in Iraq and Syria, analysts say he will need to build a broad-based coalition of international and regional players to support those efforts"

Q: What are the arguments of the trigger "combat" (event type: "Conflict:Attack")?
A: ```\nArguments = ["militants"]\n```
\end{lstlisting}
    \caption{Argument Extraction – Single instruction example}
    \label{fig:arg-extraction-single-example}
\end{figure}

\begin{table*}[t]
  \centering\small
  \setlength{\tabcolsep}{4pt}
  \begin{tabular}{lrrr}
    \toprule
    \textbf{Task Variant}                                        & \textbf{Casie} & \textbf{M2E2} & \textbf{MLEE} \\
    \midrule
    Full-Structure Construction                                  & 1,047  & 640   & 199   \\
    Role-Ablated Construction                                    &   930  & 606   & 194   \\
    Trigger Detection Only                                       & 1,047  & 640   & 199   \\
    Trigger Type Classification (Single)                         & 5,181  & 736   & 1,793 \\
    Trigger Type Classification (Multi)                          & 1,047  & 640   & 199   \\
    Trigger vs.\ Non-Trigger Discrimination (Multi)         &   931  & 526   & 193   \\
    Trigger vs.\ Non-Trigger Discrimination (Single)        & 4,184  & 1,381 & 938   \\
    Event Detection (Joint)                                      & 1,047  & 640   & 199   \\
    Argument Extraction (Single Trigger)                         & 5,183  & 736   & 1,839 \\
    Argument Extraction (Multi Triggers)                         & 1,047  & 640   & 199   \\
    Argument Extraction (Joint)                                  & 1,047  & 640   & 199   \\
    Role Assignment (Single Argument)                            & 15,466 & 1,108 & 2,760 \\
    Role Assignment (Multi Arguments)                            & 5,980  & 748   & 4,705 \\
    \midrule
    \textbf{Total}                                               & \textbf{44,137} & \textbf{9,681} & \textbf{13,616} \\
    \bottomrule
  \end{tabular}
  \caption{Number of examples per decomposed instruction variant and dataset.}
  \label{tab:decomp-dataset-stats}
\end{table*}

\section{Training and Hyperparameter Details}
This section outlines the hardware setup and key hyperparameters used to train the RoBERTa sequence tagger and fine-tune the LLM-based components in \textsc{ARIS}.

\subsection{Infrastructure for LLM Fine-Tuning}
Experiments were conducted on a single GPU per run:
\begin{itemize}
\setlength{\itemsep}{0.5ex}   
\setlength{\parskip}{0pt}     
\setlength{\topsep}{1ex}
\item CASIE/ MLEE: NVIDIA H200 (140GB)
\item M2E2: NVIDIA A100 (80GB)
\end{itemize}
Average fine-tuning times: CASIE (8h), MLEE (3h), M2E2 ($<$1~h).

\paragraph{Training} We fine-tune two instruction models: \texttt{microsoft/Phi-3-small-8k-instruct} and \texttt{meta-llama/Llama-3.1-8B-Instruct}, each using LoRA-based parameter-efficient adaptation. Both models are trained for 2 epochs with context length 4096, a batch size of 4. For Phi-3, we apply LoRA with $r{=}32$, $\alpha{=}128$, dropout 0.05, and target modules \texttt{\{q\_proj, k\_proj, v\_proj, o\_proj, gate\_proj, down\_proj, up\_proj\}}. For LLaMA-3.1, LoRA is applied to \texttt{q\_proj} and \texttt{v\_proj} with the same rank and scaling settings.

\subsection{RoBERTa Sequence Tagger}
We use \texttt{roberta-large} as the backbone encoder for all sequence tagging experiments.
\paragraph{Training} Batch sizes and epochs per dataset are summarized in below table.

\begin{center}
\begin{tabular}{lccc}
  \toprule
  Dataset & Task & Batch Size & Epochs \\
  \midrule
  CASIE & ED / EAE & 16 / 4 & 10 / 90 \\
  MLEE  & ED / EAE & 16 / 4 & 60 / 90 \\
  M2E2  & ED / EAE & 32 / 6 & 10 / 90 \\
  \bottomrule
\end{tabular}
\end{center}

\section{Examples}
\label{sec:examples}


\definecolor{smblue}{RGB}{230,245,255}      
\definecolor{smtangerine}{RGB}{255,238,224} 
\definecolor{agreegreen}{RGB}{234,255,234}  
\definecolor{warnred}{RGB}{255,235,235}     

\begin{figure*}[h]
\centering
  \footnotesize
  \setlength{\tabcolsep}{6pt}
  \begin{tabular}{@{}p{0.28\linewidth}p{0.68\linewidth}@{}}
    \toprule
    \textbf{Input Text} &
    Moments after the revered activist was escorted through a crowd, the assassin walked towards Gandhi and, at a range of just one meter, fired his gun three times, \textcolor{red}{\textbf{killing}} the man who led India’s historic revolt against British rule. \\
    \midrule
    \rowcolor{smblue}
    \textbf{Self–MoA Triggers} & \texttt{fired}, \texttt{killing}  \\
    \rowcolor{smtangerine}
    \textbf{TagPrime Triggers} & \texttt{killing} \\
    \midrule
    \rowcolor{agreegreen}
    \textbf{Agreement Set} & \texttt{killing} \\
    \rowcolor{warnred}
    \textbf{Disagreement (high‑conf.)} & -- \\
    \rowcolor{warnred}
    \textbf{Disagreement (low‑conf.)} & \texttt{fired} \emph{(discarded)} \checkmark \\
    \rowcolor{warnred}
    \textbf{Disagreement (need reflection)} & -- \\
    \midrule
    \textbf{Final Trigger List} & \texttt{killing} \\
    \midrule
    \multicolumn{2}{@{}l}{\textbf{Argument Pipeline for "\texttt{killing}"}} \\[2pt]
    \rowcolor{smblue}
    \textbf{Self–MoA Arguments} & assassin $\xrightarrow{\textit{Agent}}$ killing;\; Gandhi $\xrightarrow{\textit{Victim}}$ killing \\
    \rowcolor{smtangerine}
    \textbf{TagPrime Arguments} & assassin $\xrightarrow{\textit{Agent}}$ killing;\; man $\xrightarrow{\textit{Victim}}$ killing \\
    \midrule
    ... \\
    \midrule
    \textbf{Final Event Representation} &
      assassin $\xrightarrow{\textit{Agent}}$ killing;\;
      Gandhi $\xrightarrow{\textit{Victim}}$ killing \\
    \midrule
    \textbf{Gold Reference} &
      assassin $\xrightarrow{\textit{Agent}}$ killing;\;
      Gandhi $\xrightarrow{\textit{Victim}}$ killing;\
      gun $\xrightarrow{\textit{Instrument}}$ killing\\
    \bottomrule
  \end{tabular}
  \caption{Illustrative walk‑through of the ARIS pipeline on an \textsc{M2E2} document. 
\emph{Step 1:} the Self–MoA ensemble suggests two triggers (\texttt{killing}, \texttt{fired}) while the TagPrime outputs (\texttt{killing}). 
\emph{Step 2:} the agreement module keeps the shared trigger \texttt{killing} and flags the disagreement \texttt{fired}. 
\emph{Step 3:} confidence filtering rejects the low‑confidence \texttt{fired}. 
\emph{Step 4:} reflection resolves argument‑level mismatches.
\emph{Outcome:} the final event representation matches gold except for the still‑missing \textit{Instrument} (\texttt{gun}), revealing an open error category.}

  \label{fig:appendix_m2e2_case}
\end{figure*}
 
\paragraph{ARIS Refinement on \textsc{M2E2} sample input}
Figure~\ref{fig:appendix_m2e2_case} illustrates the three-stage ARIS pipeline: agreement detection, confidence-based filtering, and reflection-based resolution.






\label{sec:appendix}

\end{document}